\documentclass[runningheads]{llncs}

\usepackage[T1]{fontenc}

\usepackage{graphicx,verbatim}
\usepackage{booktabs}
\usepackage{multirow}
\usepackage{amsmath}

\begin{document}

\title{Can We Simplify Slide-level Fine-tuning of Pathology Foundation Models?}

\author{Jiawen Li\inst{1,*} \and
Jiali Hu\inst{1,*} \and
Qiehe Sun\inst{1} \and 
Renao Yan\inst{4} \and 
Minxi Ouyang\inst{1} \and 
Tian Guan\inst{1} \and 
Anjia Han\inst{3} \and
Chao He\inst{2,\dag} \and
Yonghong He\inst{1,\dag}}

\authorrunning{Jiawen Li et al.}

\institute{Shenzhen International Graduate School, Tsinghua University \and
Department of Engineering Science, University of Oxford \and
Department of Pathology, The First Affiliated Hospital of Sun Yat-sen University \and
Department of Mechanical Engineering, University of Washington \\
\email{chao.he@eng.ox.ac.uk, heyh@sz.tsinghua.edu.cn}}

\maketitle  

\begin{abstract}
The emergence of foundation models in computational pathology has transformed histopathological image analysis, with whole slide imaging (WSI) diagnosis being a core application. Traditionally, weakly supervised fine-tuning via multiple instance learning (MIL) has been the primary method for adapting foundation models to WSIs. However, in this work we present a key experimental finding: a simple nonlinear mapping strategy combining mean pooling and a multilayer perceptron, called SiMLP, can effectively adapt patch-level foundation models to slide-level tasks without complex MIL-based learning. Through extensive experiments across diverse downstream tasks, we demonstrate the superior performance of SiMLP with state-of-the-art methods. For instance, on a large-scale pan-cancer classification task, SiMLP surpasses popular MIL-based methods by 3.52\%. Furthermore, SiMLP shows strong learning ability in few-shot classification and remaining highly competitive with slide-level foundation models pretrained on tens of thousands of slides. Finally, SiMLP exhibits remarkable robustness and transferability in lung cancer subtyping. Overall, our findings challenge the conventional MIL-based fine-tuning paradigm, demonstrating that a task-agnostic representation strategy alone can effectively adapt foundation models to WSI analysis. These insights offer a unique and meaningful perspective for future research in digital pathology, paving the way for more efficient and broadly applicable methodologies.


\end{abstract}

\let\thefootnote\relax\footnote{* Contributed equally. $\dag$ Corresponding authors.}
\section{Introduction}
With advancements in self-supervised learning and large-scale whole-slide digitization, foundation model-based pathology AI workflows are transforming computational pathology \cite{song2023artificial,zhang2024challenges}. Self-distillation across millions of pathology images enhances region-of-interest representation \cite{chen2024towards,xu2024whole,vorontsov2024foundation}, while contrastive learning with natural language descriptions enables multimodal pathology models to integrate semantic knowledge \cite{lu2024visual,xiang2025vision,ikezogwo2023quilt}. As foundation models evolve, their ability to generalize across clinical tasks is becoming a key driver of future pathology AI development.

Representing and analyzing gigapixel-level WSIs remains a critical challenge. Traditional visual models pretrained on low-resolution natural images struggle as WSI encoders \cite{van2021deep}. A common approach is to extract tissue-containing patches and aggregate their patch-level features for slide representation fine-tuning \cite{song2023artificial,lu2021data,li2023deeptree,yan2024shapley}. Given its alignment with clinical needs, weakly supervised methods, particularly multiple instance learning (MIL), have become a widely adopted fine-tuning strategy \cite{li2021dual,shao2021transmil,xiang2023exploring,ling2024agent}. While pathology foundation models enable direct histopathology image encoding, their adaptation to WSIs still largely depends on MIL or its variants \cite{chen2024towards,xu2024whole,vorontsov2024foundation}.

Many MIL-based fine-tuning methods integrate complex feature transformations \cite{chen2022scaling,zheng2022graph,chu2024retmil,shao2021transmil,li2024dynamic} or high-order aggregation strategies \cite{zhang2024attention,zhang2022dtfd,qiehe2024nciemil,yan2024shapley,ling2024agent}, yet their necessity in the foundation model era remains uncertain. In fact, the performance of MIL-based fine-tuning with foundation models is task-dependent \cite{xu2025multiple,vaidya2024demographic}. For example, traditional MIL has outperformed more complex methods in metastasis detection \cite{ling2024towards} and breast morphological subtyping \cite{jaume2024multistain} but underperformed in lung cancer subtyping \cite{ling2024agent}; meanwhile, compared to unsupervised strategies \cite{song2024morphological}, MIL-based fine-tuning have shown unstable generalization. Given the strong features extracted by pretrained models, the advantages of complex fine-tuning strategies may be limited. Thus, exploring simplified fine-tuning approaches could offer greater efficiency, deployment flexibility, and enhanced generalization in adapting foundation models to WSIs.

In this work, we demonstrated the feasibility of simplifying slide-level fine-tuning for foundation models through extensive experiments. Using a simple combination of task-agnostic average pooling and a non-linear MLP, termed SiMLP, we seamlessly adapted foundation models to slide-level tasks. To comprehensively evaluate SiMLP, we fine-tuned three representative foundation models on six large-scale WSI classification tasks across TCGA, CPTAC, EBRAINS \cite{roetzer2022digital}, and HEROHE \cite{conde2022herohe} cohorts, achieving state-of-the-art performance. Few-shot experiments on pan-cancer tasks from TCGA and CPTAC further confirmed its superior feature representation capabilities over MIL-based methods. To assess competitiveness against pretrained slide-level foundation models, we tested it on two challenging tasks in the BRACS cohort \cite{brancati2022bracs}, where it remained highly competitive despite other models being pretrained on tens of thousands of WSIs. Finally, transferability experiments on non-small cell lung cancer subtyping across three cohorts showed that SiMLP maintains stability with minimal standard deviation, making it well-suited for scaling to large external test cohorts.

\section{Methodology}

\subsection{Weakly supervised learning on fine-tuning slide-level tasks}
\begin{figure}[t]
    \centering
    \includegraphics[width=1\linewidth]{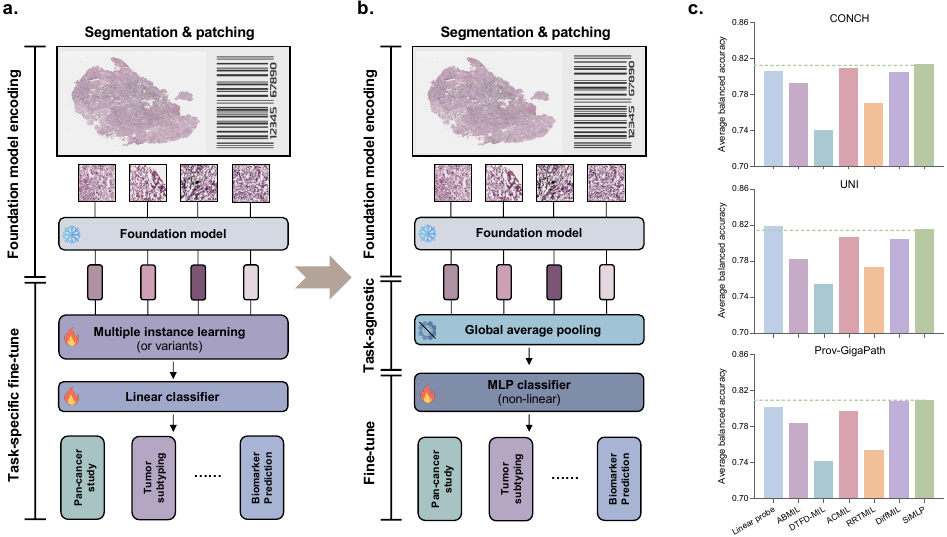}
    \caption{\textbf{Transition of slide-level adaption in pathology foundation models.} \textbf{a. }Conventional fine-tuning strategy using task-specific supervised learning. \textbf{b. }Simplified fine-tuning strategy using task-agnostic pooling and nonlinear classifier (SiMLP). \textbf{c. } Comparison of SiMLP and other MIL-based fine-tuning methods across three pathology foundation models.}
    \label{main}
\end{figure}

Before the large-scale development of pathology foundation models, visual models pretrained on natural images struggled to extract effective features from pathology images due to  their limited pathology domain understanding. Consequently, weakly supervised learning has been necessary to obtain slide representations from patch features (\textbf{Fig.\ref{main}a}). Specifically, given a WSI with patch feature set denoted as $P = \{ p_1, p_2, ..., p_n\}$, feature transformation $F(\cdot)$ and aggregation $G(\cdot)$ are proposed:
\begin{equation}
  (\tilde{p_1}, \tilde{p_2}, ..., \tilde{p_n}) = F(p_1, p_2, ..., p_n), \quad s = G(\{\tilde{p_1}, \tilde{p_2}, ..., \tilde{p_n}\}) ,
\end{equation}
where $F$ and $G$ respectively denote vector and scalar-valued functions. MIL-based fine-tuning typically follows the composition of these two functions. For instance, in the classical ABMIL \cite{ilse2018attention}, $F$ is the identity mapping and $G$ is a gated attention mechanism; whereas in TransMIL \cite{shao2021transmil}, $F$ applies a nonlinear self-attention transformation and $G$ outputs a class token. Regardless of the specific method, the composite function invariably contains learnable parameters that must be optimized using the slide-level labels from downstream tasks:
\begin{equation}
    \hat{y} =\mathrm{Softmax} (Ws),\quad L(s,y)=-\sum_{k=1}^{K}y_{k}ln(\hat{y}_k),
\end{equation}
where $W$ is a linear classifier and $L$ represents the cross-entropy loss. While effective, this approach yields task-dependent slide representations, limiting generalizability and robustness to distributional shifts.

\subsection{Slide representation with task-agnostic pooling}
Pathology foundation models pretrained over millions of histopathology images provide the possibility of obtaining task-agnostic slide representation. For instance, by clustering patch features extracted from the foundation model, WSI features can be represented as a combination of morphological prototypes \cite{song2024morphological}. Additionally, further training a slide encoder with proxy tasks based on large-scale patch features has been shown to be an effective aggregation strategy for generating generic slide-level features, both in visual \cite{xu2024whole,wang2024pathology} and multimodal \cite{jaume2024multistain,ding2024multimodal,vaidya2025molecular} settings. Although these approaches have demonstrated promising results, they often rely on additional signals for guidance. In contrast, a more straightforward approach is to leverage pooling layers, which represent one of the simplest feature aggregation methods. Pooling has been widely adopted in fine-tuning modules across various vision tasks and requires no additional learnable parameters. Therefore, the aggregation capability of pooling-based methods is worth exploring as a baseline, providing a simplified solution for slide-level fine-tuning and validating its transferability across diverse tasks.

\begin{table}[t]
\caption{\textbf{Slide-level classification on TCGA and CPTAC cohort} in terms of balanced accuracy. Best performing fine-tuning approach for each metric is bolded. 95\% CI is included in parentheses.}
\label{all_1}
\centering
\setlength\tabcolsep{6pt}   
\resizebox{1\columnwidth}{!}{
\renewcommand{\arraystretch}{1.}{
\begin{tabular}{ll|ccc}
\toprule
 & Approach & TCGA (OncoTree) & TCGA (Pan Cancer) & CPTAC (Pan Cancer) \\ 
 & & (30 classes, 2703 WSIs) & (22 classes, 2703 WSIs) & (12 classes, 1772 WSIs) \\
\midrule
\multirow{7}{*}{\rotatebox{90}{CONCH \cite{lu2024visual}}} &
Linear probe & 0.8090 (0.8032-0.8148) & 0.8702 (0.8642-0.8763) & 0.9143 (0.9126-0.9160)    \\
& ABMIL \cite{ilse2018attention} & 0.8008 (0.7877-0.8140) & 0.8539 (0.8457-0.862) & 0.8988 (0.8918-0.9059)   \\
& DTFD-MIL \cite{zhang2022dtfd} & 0.7770 (0.7720-0.7821) & 0.8423 (0.8381-0.8465) & 0.8970 (0.8922-0.9017)    \\
& ACMIL \cite{zhang2024attention} & 0.8095 (0.8021-0.8169) & 0.8618 (0.8547-0.8690) & 0.9068 (0.9033-0.9103)    \\ 
& RRTMIL \cite{tang2024feature} & 0.8221 (0.8157-0.8286) & 0.8725 (0.8584-0.8865) & 0.8116 (0.8079-0.8152)    \\ 
& DiffMIL & 0.8171 (0.8089-0.8253) & 0.8720 (0.8669-0.8771) & 0.8961 (0.8916-0.9007)    \\
& \textbf{SiMLP} & \textbf{0.8273 (0.8250-0.8295)} & \textbf{0.8788 (0.8729-0.8847)} & \textbf{0.9251 (0.9203-0.9298)}   \\
\midrule
\multirow{7}{*}{\rotatebox{90}{UNI \cite{chen2024towards}}} &
Linear probe & 0.8295 (0.8229-0.8360) & 0.8816 (0.8780-0.8851) & 0.8997 (0.8965-0.9029)    \\
& ABMIL \cite{ilse2018attention} & 0.7906 (0.7842-0.7970) & 0.8541 (0.8484-0.8598) & 0.8770 (0.8712-0.8827)   \\
& DTFD-MIL \cite{zhang2022dtfd} & 0.8127 (0.8090-0.8165) & 0.8560 (0.8495-0.8626) & 0.8595 (0.8286-0.8904)    \\
& ACMIL \cite{zhang2024attention} & 0.8240 (0.8152-0.8329) & 0.8712 (0.8651-0.8773) & 0.8968 (0.8913-0.9023)    \\ 
& RRTMIL \cite{tang2024feature} & 0.8342 (0.8198-0.8486) & 0.8720 (0.8634-0.8806) & 0.7801 (0.7713-0.7890)   \\ 
& DiffMIL & 0.8346 (0.8318-0.8374) & 0.8833 (0.8772-0.8895) & 0.8790 (0.8736-0.8844)    \\
& \textbf{SiMLP} & \textbf{0.8488 (0.8440-0.8537)} & \textbf{0.8846 (0.8821-0.8872)} & \textbf{0.9147 (0.9117-0.9176)}   \\
\midrule
\multirow{7}{*}{\rotatebox{90}{Prov-GigaPath \cite{xu2024whole}}} &
Linear probe & 0.8039 (0.7991-0.8087) & 0.8674 (0.8584-0.8764) & 0.8959 (0.8885-0.9034)    \\
& ABMIL \cite{ilse2018attention} & 0.7738 (0.7669-0.7807) & 0.8389 (0.8302-0.8475) & 0.8837 (0.8804-0.8869)   \\
& DTFD-MIL \cite{zhang2022dtfd} & 0.7852 (0.7780-0.7924) & 0.8352 (0.8256-0.8449) & 0.8827 (0.8810-0.8844)    \\
& ACMIL \cite{zhang2024attention} & 0.7996 (0.7936-0.8056) & 0.8600 (0.8478-0.8721) & 0.8947 (0.8859-0.9035)    \\
& RRTMIL \cite{tang2024feature} & 0.8147 (0.8062-0.8233) & 0.8368 (0.8253-0.8483) & 0.7849 (0.7788-0.7909)    \\ 
& DiffMIL & 0.8237 (0.8167-0.8306) & 0.8650 (0.8599-0.8701) & 0.8814 (0.8780-0.8849)    \\
& \textbf{SiMLP} & \textbf{0.8247 (0.8190-0.8304)} & \textbf{0.8739 (0.8643-0.8835)} & \textbf{0.9109 (0.9053-0.9164)}   \\ 
\bottomrule
\end{tabular}}}
\end{table}

\subsection{Non-linear classification head}
Using linear probe, a simple linear transformation, general-purpose slide representations can be widely adapted to various WSI-based clinical tasks. However, its linear nature limits its ability to effectively align representations with the lower-dimensional space of downstream tasks. To enhance the transferability of slide representations, we adopt a non-linear classifier based on a two-layer MLP. Notably, modern deep learning frameworks can efficiently optimize matrix multiplications and the additional activation layer, this adjustment strikes a balance between improving representation flexibility and maintaining efficiency. The overall of SiMLP is shown in \textbf{Fig.\ref{main}b}.

\begin{table}[t]
\caption{\textbf{Slide-level classification on EBRAINS and HEROHE cohort} in terms of balanced accuracy. Best performing fine-tuning approach for each metric is bolded. 95\% CI is included in parentheses.}
\label{all_2}
\centering
\setlength\tabcolsep{6pt}   
\resizebox{1\columnwidth}{!}{
\renewcommand{\arraystretch}{1.}{
\begin{tabular}{ll|ccc}
\toprule
& Approach & EBRAINS (Subtyping) & EBRAINS (\textit{IDH} Pred.) & HEROHE (\textit{HER2} Pred.) \\ 
& & (27 classes, 649 WSIs) & (2 classes, 208 WSIs) & (2 classes, 149 WSIs) \\
\midrule
\multirow{7}{*}{\rotatebox{90}{CONCH \cite{lu2024visual}}} &
Linear probe & 0.6391 (0.6312-0.6471) & 0.8456 (0.8281-0.8632) & \textbf{0.7578 (0.7493-0.7663)}    \\
& ABMIL \cite{ilse2018attention} & 0.6366 (0.6283-0.6449) & 0.8398 (0.8246-0.8549) & 0.7268 (0.6885-0.7652)   \\
& DTFD-MIL \cite{zhang2022dtfd} & 0.5323 (0.5251-0.5395) & 0.6871 (0.6649-0.7093) & 0.7036 (0.6707-0.7365)   \\
& ACMIL \cite{zhang2024attention} & 0.6620 (0.6474-0.6766) & \textbf{0.8650 (0.8584-0.8716)} & 0.7518 (0.7438-0.7598)    \\ 
& RRTMIL \cite{tang2024feature} & 0.6084 (0.5875-0.6292) & 0.8325 (0.8126-0.8524) & 0.6770 (0.6379-0.7162)    \\ 
& DiffMIL & 0.6628 (0.6411-0.6844) & 0.8388 (0.8293-0.8482) & 0.7433 (0.7227-0.7639)    \\
& \textbf{SiMLP} & \textbf{0.6763 (0.6641-0.6884)} & 0.8567 (0.8417-0.8716) & 0.7149 (0.6720-0.7578)   \\
\midrule
\multirow{7}{*}{\rotatebox{90}{UNI \cite{chen2024towards}}} &
Linear probe & 0.6818 (0.6728-0.6908) & \textbf{0.8879 (0.8797-0.8961)} & \textbf{0.7325 (0.7144-0.7507)}    \\
& ABMIL \cite{ilse2018attention} & 0.6501 (0.6312-0.6690) & 0.8371 (0.8069-0.8672) & 0.6829 (0.6498-0.7161)   \\
& DTFD-MIL \cite{zhang2022dtfd} & 0.5843 (0.5737-0.5949) & 0.7058 (0.6884-0.7233) & 0.7090 (0.6874-0.7307)    \\
& ACMIL \cite{zhang2024attention} & 0.6873 (0.6713-0.7032) & 0.8671 (0.8578-0.8764) & 0.6920 (0.6403-0.7437)    \\ 
& RRTMIL \cite{tang2024feature} & 0.6189 (0.6035-0.6342) & 0.8565 (0.8495-0.8634) & 0.6776 (0.6449-0.7103)   \\ 
& DiffMIL & 0.6850 (0.6694-0.7006) & 0.8263 (0.8173-0.8352) & 0.7193 (0.7084-0.7302)    \\
& \textbf{SiMLP} & \textbf{0.6940 (0.6865-0.7014)} & 0.8790 (0.8660-0.8919) & 0.6703 (0.6348-0.7057)   \\
\midrule
\multirow{7}{*}{\rotatebox{90}{Prov-GigaPath \cite{xu2024whole}}} &
Linear probe & 0.6915 (0.6859-0.6971) & 0.8567 (0.8395-0.8739) & 0.6913 (0.6762-0.7064)    \\
& ABMIL \cite{ilse2018attention} & 0.6717 (0.6593-0.6841) & 0.8492 (0.8390-0.8594) & 0.6869 (0.6312-0.7425)
   \\
& DTFD-MIL \cite{zhang2022dtfd} & 0.5456 (0.4963-0.5949) & 0.7692 (0.7190-0.8193) & 0.6305 (0.6135-0.6476)     \\
& ACMIL \cite{zhang2024attention} & 0.7069 (0.6916-0.7221) & 0.8646 (0.8508-0.8785) & 0.6576 (0.6326-0.6826)    \\
& RRTMIL \cite{tang2024feature} & 0.6171 (0.6050-0.6291) & 0.8290 (0.8085-0.8496) & 0.6368 (0.5943-0.6794)    \\ 
& DiffMIL & \textbf{0.7161 (0.7037-0.7285)} & 0.8570 (0.8352-0.8788) &\textbf{ 0.7092 (0.6800-0.7384)}
    \\
& \textbf{SiMLP} & 0.6978 (0.6841-0.7116) & \textbf{0.8726 (0.8545-0.8906)} & 0.6778 (0.6487-0.7069)   \\ 
\bottomrule
\end{tabular}}}
\end{table}

\section{Experiments}

\subsection{Datasets and experimental settings}
We conducted extensive experiments on seven large-scale datasets, including \textbf{TCGA} (OncoTree, 30 classes), \textbf{TCGA} (Pan Cancer, 22 classes), \textbf{CPTAC} (Pan Cancer, 12 classes),  \textbf{EBRAINS} (Subtyping, 27 classes),  \textbf{EBRAINS} (\textit{IDH} Prediction, 2 classes), \textbf{HEROHE} (\textit{HER2} Prediction, 2 classes) and \textbf{BRACS} (Coarse-grained, 3 classes; Fine-grained 7 classes). We trained all the fine-tuning approaches with AdamW optimizer (learning rate: $10^{-4}$, betas=[0.9, 0.98], weight decay: $10^{-4}$) and a batch size of 1 for 20 epochs. All approaches were trained on $1\times24$GB NVIDIA 4090 with 5 fixed random seeds. Additional details of implementation, datasets, and baselines will be available in the Github codebase.

\subsection{SiMLP outperforms in diverse slide-level classification}
To evaluate SiMLP across slide-level tasks, we selected three representative pathology foundation models: CONCH \cite{lu2024visual}, UNI \cite{chen2024towards}, and Prov-GigaPath \cite{xu2024whole}. We conducted experiments on six tasks across four cohorts and performed a fair comparison against linear probe, four popular MIL-based methods (ABMIL \cite{ilse2018attention}, DTFD-MIL \cite{zhang2022dtfd}, ACMIL \cite{zhang2024attention}, and RRT-MIL \cite{tang2024feature}), and a differential attention-based MIL method (DiffMIL) that we specifically designed (\textbf{Table \ref{all_1}-\ref{all_2}}). Overall, SiMLP achieved superior performance across all three foundation models (81.32\%, 81.52\%, 80.96\% in \textbf{Fig.\ref{main}c}), demonstrating stronger adaptability than task-specific weakly supervised learning. Notably, SiMLP achieved the best results in three pan-cancer tasks, improving upon ABMIL by 3.52\% and ACMIL by 1.83\% in TCGA OncoTree classification. While SiMLP underperformed in \textit{HER2} prediction, the linear probe, which also uses mean pooling, performed well, suggesting that task-agnostic simplified aggregation can still produce effective representations.

\begin{figure}[t]
    \centering
    \includegraphics[width=0.9\linewidth]{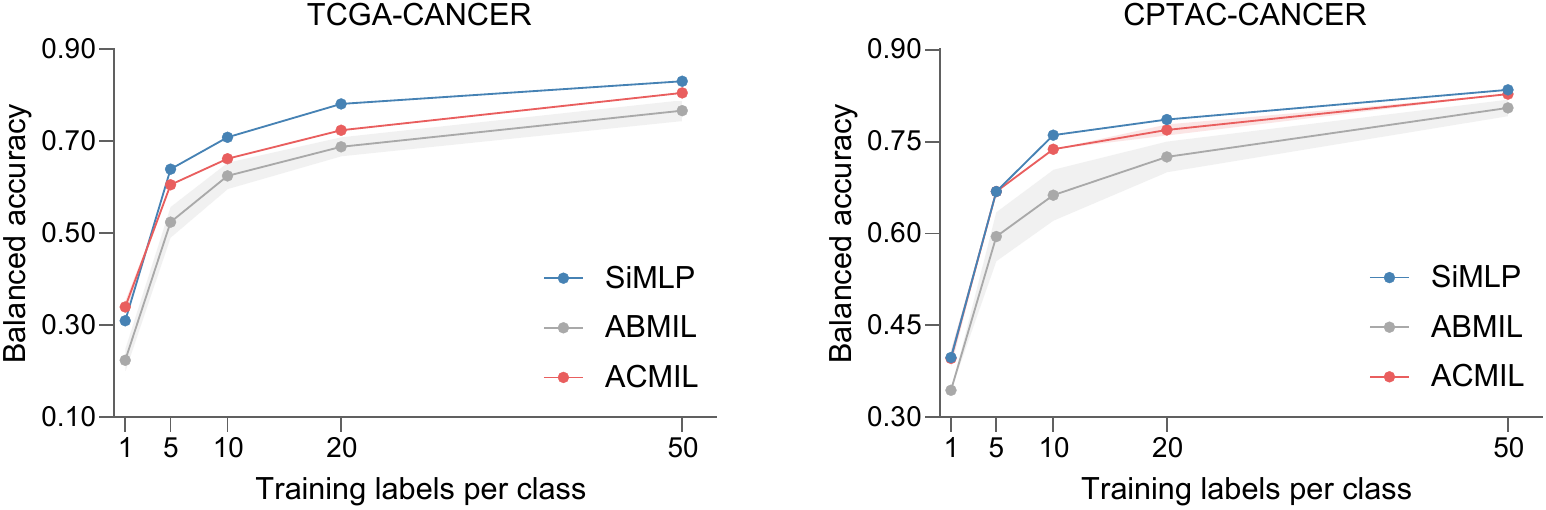}
    \caption{\textbf{Few-shot slide-level performance on TCGA and CPTAC cohort} with $K\in\{1,5,10,20,50\}$ slides per class.}
    \label{few_shot}
\end{figure}

\begin{table}[t]
\caption{\textbf{Comparison with slide-level foundation models on BRACS cohort} in terms of balanced accuracy, ROC AUC, and weighted F1 score (reported as averages). Best performing approach for each metric is bolded.}
\label{slide_level}
\centering
\setlength\tabcolsep{6pt}   
\resizebox{1\columnwidth}{!}{
\renewcommand{\arraystretch}{1.2}{
\begin{tabular}{l|ccc|ccc}
\toprule
Approach & \multicolumn{3}{c}{BRACS (Coarse-grained, 3 classes)} & \multicolumn{3}{|c}{BRACS (Fine-grained, 7 classes)} \\
 & Bal ACC     & ROC AUC  & Weighted F1 & Bal ACC  & ROC AUC     & Weighted F1 \\
\midrule
CHIEF \cite{wang2024pathology} \textit{with} Linear probe    & 0.5438 & 0.8195  & 0.5089    & 0.2732 & 0.7353 & 0.2506    \\
CHIEF \cite{wang2024pathology} \textit{with} Full tuning     & \textbf{0.5833} & \textbf{0.8249} & \textbf{0.5457}    & \textbf{0.2780} & \textbf{0.7663} & 0.2665     \\
\textbf{SiMLP} \textit{with} CTransPath \cite{wang2022transformer}     & 0.5155 & 0.7433 & 0.5250    & 0.2518 & 0.6534  & \textbf{0.2955}    \\
\midrule
GigaPath \cite{xu2024whole} \textit{with} Linear probe  & 0.3771 & 0.7298 & 0.3220 & 0.2289 & 0.6757 & 0.2393     \\
GigaPath \cite{xu2024whole} \textit{with} Full tuning  & 0.3333 & 0.4409 & 0.1978    & 0.1429 & 0.5047 & 0.0335    \\
\textbf{SiMLP} \textit{with} Prov-GigaPath \cite{xu2024whole}  & \textbf{0.5409} & \textbf{0.7419} & \textbf{0.5474}    & \textbf{0.2516} & \textbf{0.6772} & \textbf{0.2959}   \\
\bottomrule
\end{tabular}}}
\end{table}

\subsection{SiMLP outperforms in few-shot learning classification}
To evaluate learning efficiency and generalization with limited data, we conducted few-shot classification on TCGA and CPTAC pan-cancer tasks using UNI (\textbf{Fig.\ref{few_shot}}). We trained SiMLP, ABMIL, and ACMIL with $K\in\{1,5,10,20,50\}$ samples per class. The results show that SiMLP consistently outperformed other methods across nearly all shot settings while exhibiting lower variance across random seeds ($\mathrm{std.} < 0.01$ per shot). These results highlight that SiMLP has potential for screening rare and underrepresented clinical conditions.

\begin{figure}[t]
    \centering
    \includegraphics[width=1\linewidth]{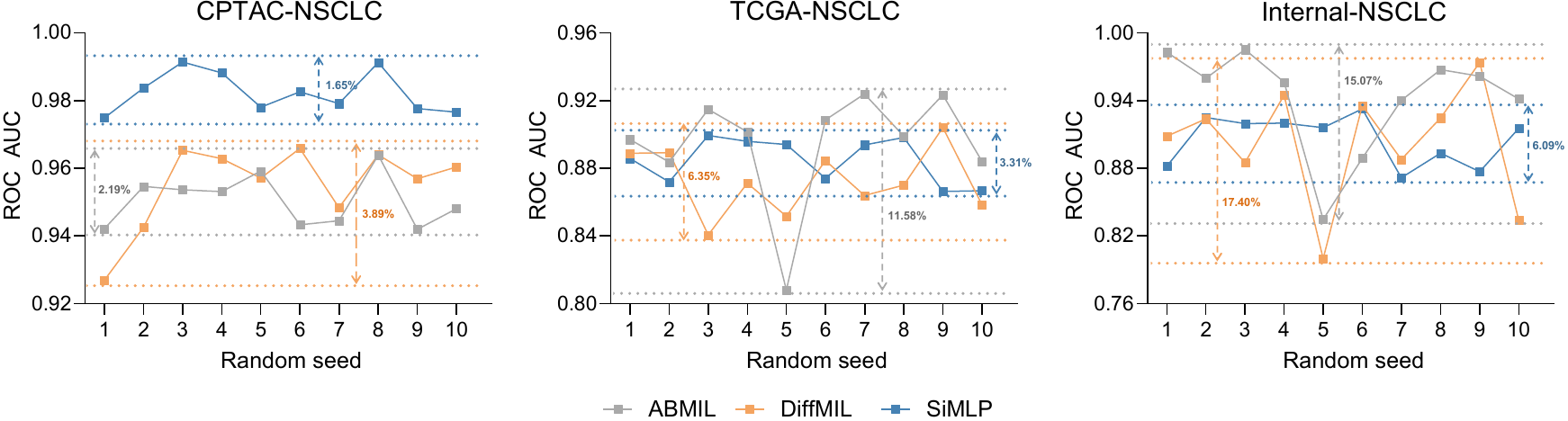}
    \caption{\textbf{Robustness and transfer testing evaluation on CPTAC, TCGA, and in-house NSCLC cohort} by sweeping 10 random seeds.} 
    \label{transfer}
\end{figure}

\begin{table}[t]
\caption{\textbf{Ablation study with on pooling and activation functions on TCGA-OncoTree} in tems of balanced accuracy, ROC AUC, and weighted F1 score (reported as averages). Best performing configuration for each metric is bolded.}
\label{ablation}
\centering
\setlength\tabcolsep{6pt}   
\resizebox{1\columnwidth}{!}{
\renewcommand{\arraystretch}{1.2}{
\begin{tabular}{l|ccc|l|ccc}
\toprule
Configuration                  & Bal ACC & ROC AUC & Weighted F1 & Configuration         & Bal ACC & ROC AUC & Weighted F1 \\
\midrule
Mean + ReLU & 0.8488  & 0.9946  & 0.8893  & Max + ReLU & 0.7456 & 0.9927 & 0.8173 \\
Mean + GeLU & \textbf{0.8509}  & \textbf{0.9949} & \textbf{0.8901} & Max + GeLU   & 0.7140 & 0.9920 & 0.8027 \\
Mean + SwigLU & 0.8054 & 0.9833 & 0.8533 & Max + SwigLU & 0.5871 & 0.9645 & 0.7049 \\
\bottomrule
\end{tabular}}}
\end{table}

\subsection{SiMLP is competitive with slide-level foundation models}
We compared SiMLP with two pretrained slide-level foundation models, CHIEF \cite{wang2024pathology} and GigaPath \cite{xu2024whole} (\textbf{Table \ref{slide_level}}), using the BRACS cohort, a challenging breast cancer subtype classification dataset with coarse-grained (3-class) and fine-grained (7-class) tasks. CHIEF employs CTransPath \cite{wang2022transformer} as its patch feature extractor, while GigaPath uses Prov-GigaPath. For fair comparison, we evaluated SiMLP under the same patch-level foundation model, applying both linear probing and full parameter fine-tuning. Results show that while SiMLP underperforms CHIEF overall, it achieves higher weighted F1 scores in fine-grained classification. Compared to GigaPath, SiMLP outperforms across all metrics in both tasks, likely due to the high computational complexity and large parameter size of GigaPath, which may hinder convergence during downstream fine-tuning. Given that CHIEF and GigaPath were pretrained on tens of thousands of WSIs, the competitive performance of SiMLP is particularly noteworthy.

\subsection{SiMLP has a good transfer capability}
We further evaluated the transferability across cohorts by constructing an NSCLC subtype classification task using LSCC and LUAD cases from CPTAC, TCGA, and an in-house (IH-LUNG) cohort. We used UNI to train ABMIL, DiffMIL, and SiMLP on CPTAC with 10 random seeds, followed by CPTAC internal testing and TCGA, IH-LUNG external testing (\textbf{Fig.\ref{transfer}}). Results show that SiMLP outperforms other methods in internal testing and exhibits greater stability than both baselines in external test sets. This highlights that SiMLP provides better generalization and robustness in transfer learning scenarios.

\subsection{Ablation study}
Finally, we conducted an ablation study on SiMLP. Specifically, we replaced mean pooling with max pooling and examined the effect of substituting the ReLU activation function with GeLU and SwigLU in different combinations. These modifications were evaluated on the TCGA-OncoTree task with UNI encoder (\textbf{Table \ref{ablation}}). The results show that slide representations generated using max pooling perform worse than those generated with mean pooling, indicating that capturing global features remains crucial for task-agnostic aggregation. Additionally, we observed that the combination of GeLU and mean pooling led to improved performance, suggesting that adjusting the non-linear classifier further enhances adaptation to downstream tasks.

\section{Conclusion and future direction}
In this work we found that SiMLP, a simple fine-tuning method, enables pathology foundation models to effectively adapt to slide-level tasks. Extensive experiments demonstrate that SiMLP outperforms widely used MIL-based weakly supervised learning, confirming its strong performance and generalization ability.

Our findings provide four key insights for the future of computational pathology in the foundation model era:

1. \textbf{Patch-level foundation model development remains crucial.} While existing pretrained encoders enhance WSI analysis, balancing data redundancy and model complexity is essential. For instance, ViT-Base (CONCH) performed competitively against ViT-Giant (Prov-GigaPath). We encourage future research to explore efficient architectures, diverse multimodal models, and improved data-driven preprocessing strategies.

2. \textbf{Task-agnostic slide representation learning may be more impactful than weakly supervised learning.} Such representations improve generalization and stability while enabling broader applications like slide embedding retrieval and convenient multimodal integration.

3. \textbf{Advancing slide-level foundation models enhances clinical performance.} Pretraining slide encoders on large-scale datasets not only supports task-agnostic representation learning but also allows for performance improvements through diverse fine-tuning strategies.

4. \textbf{Tailored weakly supervised learning remains necessary for slide-level tasks.} SiMLP performs well broadly, however, weakly supervised learning still holds advantages in specific tasks, highlighting its effectiveness for clinically tailored applications. For example, it remains valuable for biomarker prediction, hierarchical classification of rare diseases \cite{jin2024hmil,li2024diagnostic}, and long-tailed data analysis.

In summary, as pathology foundation models continue to evolve, simplifying traditional weakly supervised learning paradigms and pioneering a new generation of research directions will be key to further enhancing performance and enabling broader real-world applications in computational pathology.

\section{Acknowledgement}
This work was supported by the National Natural Science Foundation of China (NSFC) under Grant No.82430062, the Shenzhen Engineering Research Centre under Grant XMHT20230115004. We thank the Jilin
FuyuanGuan Food Group Co., Ltd for their collaboration. C.H. was also supported by the St John's College, the University of Oxford, and the Royal Society (URF\textbackslash R1\textbackslash 241734). The authors have no competing interests to declare
that are relevant to the content of this paper.

\bibliographystyle{splncs04}
\bibliography{ref}

\begin{thebibliography}{10}
\providecommand{\url}[1]{\texttt{#1}}
\providecommand{\urlprefix}{URL }
\providecommand{\doi}[1]{https://doi.org/#1}

\bibitem{brancati2022bracs}
Brancati, N., Anniciello, A.M., Pati, P., et~al.: Bracs: A dataset for breast carcinoma subtyping in h\&e histology images. Database  \textbf{2022},  baac093 (2022)

\bibitem{chen2022scaling}
Chen, R.J., Chen, C., Li, Y., et~al.: Scaling vision transformers to gigapixel images via hierarchical self-supervised learning. In: CVPR. pp. 16144--16155 (2022)

\bibitem{chen2024towards}
Chen, R.J., Ding, T., Lu, M.Y., et~al.: Towards a general-purpose foundation model for computational pathology. Nature Medicine  \textbf{30}(3),  850--862 (2024)

\bibitem{chu2024retmil}
Chu, H., Sun, Q., Li, J., et~al.: Retmil: Retentive multiple instance learning for histopathological whole slide image classification. In: MICCAI. pp. 437--447. Springer (2024)

\bibitem{conde2022herohe}
Conde-Sousa, E., Vale, J., Feng, M., et~al.: Herohe challenge: predicting her2 status in breast cancer from hematoxylin--eosin whole-slide imaging. Journal of Imaging  \textbf{8}(8), ~213 (2022)

\bibitem{ding2024multimodal}
Ding, T., Wagner, S.J., Song, A.H., et~al.: Multimodal whole slide foundation model for pathology. arXiv preprint arXiv:2411.19666  (2024)

\bibitem{ikezogwo2023quilt}
Ikezogwo, W., Seyfioglu, S., Ghezloo, F., et~al.: Quilt-1m: One million image-text pairs for histopathology. NeurIPS  \textbf{36},  37995--38017 (2023)

\bibitem{ilse2018attention}
Ilse, M., Tomczak, J., Welling, M.: Attention-based deep multiple instance learning. In: ICML. pp. 2127--2136. PMLR (2018)

\bibitem{jaume2024multistain}
Jaume, G., Vaidya, A., Zhang, A., et~al.: Multistain pretraining for slide representation learning in pathology. In: ECCV. pp. 19--37. Springer (2024)

\bibitem{jin2024hmil}
Jin, C., Luo, L., Lin, H., et~al.: Hmil: Hierarchical multi-instance learning for fine-grained whole slide image classification. IEEE Transactions on Medical Imaging  (2024)

\bibitem{van2021deep}
Van~der Laak, J., Litjens, G., Ciompi, F.: Deep learning in histopathology: the path to the clinic. Nature medicine  \textbf{27}(5),  775--784 (2021)

\bibitem{li2021dual}
Li, B., Li, Y., Eliceiri, K.W.: Dual-stream multiple instance learning network for whole slide image classification with self-supervised contrastive learning. In: CVPR. pp. 14318--14328 (2021)

\bibitem{li2024dynamic}
Li, J., Chen, Y., Chu, H., et~al.: Dynamic graph representation with knowledge-aware attention for histopathology whole slide image analysis. In: CVPR. pp. 11323--11332 (2024)

\bibitem{li2023deeptree}
Li, J., Cheng, J., Meng, L., et~al.: Deeptree: Pathological image classification through imitating tree-like strategies of pathologists. IEEE Transactions on Medical Imaging  \textbf{43}(4),  1501--1512 (2023)

\bibitem{li2024diagnostic}
Li, J., Sun, Q., Yan, R., et~al.: Diagnostic text-guided representation learning in hierarchical classification for pathological whole slide image. arXiv preprint arXiv:2411.10709  (2024)

\bibitem{ling2024towards}
Ling, X., Lei, Y., Li, J., et~al.: Towards a comprehensive benchmark for pathological lymph node metastasis in breast cancer sections. arXiv preprint arXiv:2411.10752  (2024)

\bibitem{ling2024agent}
Ling, X., Ouyang, M., Wang, Y., et~al.: Agent aggregator with mask denoise mechanism for histopathology whole slide image analysis. In: ACM Multimedia. pp. 2795--2803 (2024)

\bibitem{lu2024visual}
Lu, M.Y., Chen, B., Williamson, D.F., et~al.: A visual-language foundation model for computational pathology. Nature Medicine  \textbf{30}(3),  863--874 (2024)

\bibitem{lu2021data}
Lu, M.Y., Williamson, D.F., Chen, T.Y., et~al.: Data-efficient and weakly supervised computational pathology on whole-slide images. Nature Biomedical Engineering  \textbf{5}(6),  555--570 (2021)

\bibitem{roetzer2022digital}
Roetzer-Pejrimovsky, T., Moser, A.C., Atli, B., et~al.: The digital brain tumour atlas, an open histopathology resource. Scientific Data  \textbf{9}(1), ~55 (2022)

\bibitem{shao2021transmil}
Shao, Z., Bian, H., Chen, Y., et~al.: Transmil: Transformer based correlated multiple instance learning for whole slide image classification. NeurIPS  \textbf{34},  2136--2147 (2021)

\bibitem{song2024morphological}
Song, A.H., Chen, R.J., Ding, T., et~al.: Morphological prototyping for unsupervised slide representation learning in computational pathology. In: CVPR. pp. 11566--11578 (2024)

\bibitem{song2023artificial}
Song, A.H., Jaume, G., Williamson, D.F., et~al.: Artificial intelligence for digital and computational pathology. Nature Reviews Bioengineering  \textbf{1}(12),  930--949 (2023)

\bibitem{qiehe2024nciemil}
Sun, Q., Jiang, D., Li, J., Yan, R., He, Y., Guan, T., Cheng, Z.: Nciemil: Rethinking decoupled multiple instance learning framework for histopathological slide classification. In: MIDL (2024)

\bibitem{tang2024feature}
Tang, W., Zhou, F., Huang, S., et~al.: Feature re-embedding: Towards foundation model-level performance in computational pathology. In: CVPR. pp. 11343--11352 (2024)

\bibitem{vaidya2024demographic}
Vaidya, A., Chen, R.J., Williamson, D.F., et~al.: Demographic bias in misdiagnosis by computational pathology models. Nature Medicine  \textbf{30}(4),  1174--1190 (2024)

\bibitem{vaidya2025molecular}
Vaidya, A., Zhang, A., Jaume, G., et~al.: Molecular-driven foundation model for oncologic pathology. arXiv preprint arXiv:2501.16652  (2025)

\bibitem{vorontsov2024foundation}
Vorontsov, E., Bozkurt, A., Casson, A., et~al.: A foundation model for clinical-grade computational pathology and rare cancers detection. Nature Medicine  \textbf{30}(10),  2924--2935 (2024)

\bibitem{wang2022transformer}
Wang, X., Yang, S., Zhang, J., et~al.: Transformer-based unsupervised contrastive learning for histopathological image classification. Medical Image Analysis  \textbf{81},  102559 (2022)

\bibitem{wang2024pathology}
Wang, X., Zhao, J., Marostica, E., et~al.: A pathology foundation model for cancer diagnosis and prognosis prediction. Nature  \textbf{634}(8035),  970--978 (2024)

\bibitem{xiang2025vision}
Xiang, J., Wang, X., Zhang, X., et~al.: A vision--language foundation model for precision oncology. Nature pp. 1--10 (2025)

\bibitem{xiang2023exploring}
Xiang, J., Zhang, J.: Exploring low-rank property in multiple instance learning for whole slide image classification. In: ICLR (2023)

\bibitem{xu2024whole}
Xu, H., Usuyama, N., Bagga, J., et~al.: A whole-slide foundation model for digital pathology from real-world data. Nature pp.~1--8 (2024)

\bibitem{xu2025multiple}
Xu, H., Wang, M., Shi, D., et~al.: When multiple instance learning meets foundation models: advancing histological whole slide image analysis. Medical Image Analysis  \textbf{101},  103456 (2025)

\bibitem{yan2024shapley}
Yan, R., Sun, Q., Jin, C., et~al.: Shapley values-enabled progressive pseudo bag augmentation for whole-slide image classification. IEEE Transactions on Medical Imaging  (2024)

\bibitem{zhang2022dtfd}
Zhang, H., Meng, Y., Zhao, Y., et~al.: Dtfd-mil: Double-tier feature distillation multiple instance learning for histopathology whole slide image classification. In: CVPR. pp. 18802--18812 (2022)

\bibitem{zhang2024challenges}
Zhang, S., Metaxas, D.: On the challenges and perspectives of foundation models for medical image analysis. Medical Image Analysis  \textbf{91},  102996 (2024)

\bibitem{zhang2024attention}
Zhang, Y., Li, H., Sun, Y., et~al.: Attention-challenging multiple instance learning for whole slide image classification. In: ECCV. pp. 125--143. Springer (2024)

\bibitem{zheng2022graph}
Zheng, Y., Gindra, R.H., Green, E.J., et~al.: A graph-transformer for whole slide image classification. IEEE Transactions on Medical Imaging  \textbf{41}(11),  3003--3015 (2022)

\end{thebibliography}

\end{document}